\documentclass{article}
\usepackage{amsfonts,amssymb,amsmath,amsthm}
\usepackage{geometry}
\usepackage{nopageno}
\usepackage{fancyhdr}
\usepackage{graphicx}
\usepackage[numbers]{natbib}
\usepackage{subfig}
\usepackage{lastpage}
\usepackage{url}
\usepackage{float}
\usepackage{color}

\DeclareMathOperator*{\argmin}{\arg\!\min}

\begin{document}

\title{Improved Clustering with Augmented k-means}

\author{J. Andrew Howe, ahowe42@gmail.com\\
Independent Researcher, Riyadh, Saudi Arabia}
\date{}
\maketitle

\begin{abstract}
Identifying a set of homogeneous clusters in a heterogeneous dataset is one of the most important classes of problems in statistical modeling. In the realm of unsupervised partitional clustering, k-means is a very important algorithm for this. In this technical report, we develop a new k-means variant called Augmented k-means, which is a hybrid of k-means and logistic regression. During each iteration, logistic regression is used to predict the current cluster labels, and the cluster belonging probabilities are used to control the subsequent re-estimation of cluster means. Observations which can't be firmly identified into clusters are excluded from the re-estimation step. This can be valuable when the data exhibit many characteristics of real datasets such as heterogeneity, non-sphericity, substantial overlap, and high scatter. Augmented k-means frequently outperforms k-means by more accurately classifying observations into known clusters and / or converging in fewer iterations. We demonstrate this on both simulated and real datasets. Our algorithm is implemented in Python and will be available with this report. 

\end{abstract}

\section{\label{sec_intro}Introduction}
Clustering $n$ datapoints in $p$ dimensions into $K$ distinct clusters is a very old problem in statistical modeling. The k-means algorithm, introduced by MacQueen \cite{MacQueen1967}, is an important method\footnote{As of a 2002 survey \cite{Berkhin2002}.} to do this. Fundamentally, the k-means algorithm iterates through a set of steps in an attempt to minimize the sum of squared distances within all $K$ clusters. Its popularity is probably due to this inherent simplicity, as opposed to its perfection. Indeed, as mentioned in Krishna and Murty \cite{KrishnaMurty1999}, the k-means algorithm exhibits a strong tendency to converge to suboptimal local minima, and is not robust to the initial state, leading to different ways to cluster the same dataset. Many researchers have proposed differing degrees of variations around the underlying theme of iteratively minimizing the total sum of squared distances.

Wong \cite{Wong1982} developed a hybrid clustering algorithm using both k-means and single-linkage hierarchical clustering, with the specific goal of identifying high-density modal regions. With a similar goal of finding tight, stable clusters, Tseng and Wong \cite{TsengWong2005} used a re-sampling method with a merged k-means and truncated hierarchical clustering algorithm. To combat the same issue of scattered observations which truly don't belong in any cluster, Maitra and Ramler \cite{MaitraRamler2009} proposed a new algorithm which iteratively builds each homogenous spherical cluster around a core, so that scattered observations are explicitly excluded.

Several authors have used trimming in clustering with k-means, in which certain observations are remove (or trimmed). The goal of trimming is to identify clusters robustly w.r.t. outliers, originally proposed by Gordaliza \cite{Gordaliza1991}. Cuesta-Albertos \textit{et al.} used a data-based trimming method optimized by simulated annealing \cite{Cuesta-Albertosetal1997}. Garc\'{i}a-Escudero \textit{et al.} \cite{Garcia-Escuderoetal2008} introduced a trimmed clustering algorithm, constrained by the ratio of maximum to minimum eigenvalues from the within-cluster scatter matrices. Garc\'{i}a-Escuder \textit{et al.} \cite{Garcia-Escuderoetal2009} further extended robust clustering with trimmed k-means by modeling linear patterns in the data around which clusters formed.

Bozdogan \cite{Bozdogan1983} proposed an initialization method that initializes the clusters so that they are evenly spaced throughout the data. Arthur \& Vassilvitskii \cite{ArthurVassilvitskii2007} proposed a modified algorithm called \emph{k-means++}, wherein cluster centers are spaced around each other following an iterative distance-weighting scheme.

Krishna and Murty \cite{KrishnaMurty1999} created a hybrid algorithm called \emph{Genetic k-means} based on the Genetic Algorithm of Holland \cite{Holland1975,Holland1992}. Along similar lines, Song \textit{et al.} created their \textit{GARM} algorithm, which computes the cluster distances using a regularized Mahalanobis distance. Use of the Genetic Algorithm in both cases allows the clustering algorithm to better avoid local optima and robustifies it against initialization.

Perhaps most conceptually relevant to this article is the work by Tibshirani and Walther \cite{TibshiraniWalther2005}. Like \cite{TsengWong2005,Wong1982}, they focused on identifying stable clusters. Their approach used iterated k-fold cross-validation to create a hybrid unsupervised and supervised clustering prediction technique.

In this work, we propose a new variant called \textit{Augmented k-means} in which each iteration is augmented by performing logistic regression. Hence, our algorithm joins unsupervised and supervised clustering. The cluster-belonging probabilities output by the regression are used to exclude some observations from being used to re-estimate the cluster means. While this certainly adds computation time, the augmented algorithm tends to more accurately classify the observations into known clusters, and often converges in fewer iterations.

For data with homogeneous, non-overlapping clusters and little scatter, Augmented k-means should require more time and iterations than k-means to converge to a solution, assuming the clusters are seeded well (as in \cite{ArthurVassilvitskii2007,Bozdogan1983}). However, in our experience, real datasets for which clustering is needed are often generated by diffuse, heterogenous, non-spherical, and highly overlapped populations. Hence, Augmented k-means is a practical addition to the current set of clustering methodologies.

In the interest of reproducible and open research, the Augmented k-means algorithm is implemented in Python using the scientific computing package numpy and the machine learning library scikit-learn \cite{scikit-learn}. Along with the code for running the Monte Carlo experiments, it will be available with this report.

Our new algorithm is detailed in Section~\ref{sec_algorithm}, followed by numerical results on both simulated and real datasets in Section~\ref{sec_results}. We finish with some concluding remarks in Section~\ref{sec_conclude}.

\section{\label{sec_algorithm}Augmented k-means}
The k-means algorithm typically starts with $K$ initial cluster means\footnote{Alternatively, it can start by classifying each observation into $K$ clusters, but this is a trivial difference}. From here, it iterates assigning observations into their closest cluster, and recomputing the cluster means. The notation we use is:
\begin{itemize}
\item $\mathbf{x_i}$: the $i$th observation vector, $i=1,\ldots n$
\item $y_i$: the cluster assignment of the $i$th observation, $y_i\in\left\{1,\ldots,K\right\}$, $i=1,\ldots n$
\item $I\left(i,k\right)$: returns $1$ if $y_i=k$, and returns $0$ otherwise
\item $\mathbf{C_k}$: the mean vector of the $k$th cluster, $k=1,\ldots,K$
\item $\epsilon$: convergence criteria for sequential difference in total sum of squared distances
\end{itemize}
After generating $K$ initial cluster means - we use the initialization from k-means++ \cite{ArthurVassilvitskii2007} - the k-means algorithm is:
\newpage
\begin{enumerate}
\item For each observation $\mathbf{x_i}$ and cluster $k$, compute the squared Euclidean distance to the mean $\mathbf{C_k}$:
\begin{equation}
d_{ik} = \parallel\mathbf{x_i}-\mathbf{C_k}\parallel^2_2.
\end{equation}

\item Assign each observation to its closest cluster:
\begin{equation}
y_i = \underset{k=1,\ldots,K}{\argmin}d_{ik}.
\end{equation}

\item Recompute the cluster means:
\begin{equation}
\mathbf{C_k} = \frac{1}{n_k}\sum_{i=1}^n I\left(i,k\right)\mathbf{x_i},\ 
n_k = \sum_{i=1}^n I\left(i,k\right)
\end{equation}

\item Compute the total sum of squared distances:
\begin{equation}
S_t = \sum_{k=1}^K\sum_{i=1}^n I\left(i,k\right)d_{ik}.
\end{equation}

\item Measure the change in the total sum of squared distances from the previous iteration\footnote{Only starting from the second iteration, obviously.}, and compare against $\epsilon$. If $\left\vert S_{t-1} - S_{t} \right\vert<\epsilon$, exit. Otherwise, return to step (i).
\end{enumerate}
After step (ii), we have $n$ cluster assignments $y_i$. If we take the stance that they are known class labels, we can use this data to formulate a supervised learning problem. Accordingly, Augmented k-means inserts one step after (ii) and modifies (iii). This new step begins by solving the set of multinomial\footnote{Obviously, if $K=2$, regular binary logistic regression is used.} logistic regression equations shown below.
\begin{equation}
\begin{aligned}
\Pr(y_i=K)=&{\frac {1}{1+\sum _{k=1}^{K-1}e^{{\boldsymbol {\beta }}_{k}\cdot \mathbf x_i}}}\\
\Pr(y_i=1)=&{\frac {e^{{\boldsymbol {\beta }}_{1}\cdot \mathbf x_i}}{1+\sum _{k=1}^{K-1}e^{{\boldsymbol {\beta }}_{k}\cdot \mathbf x_i}}}\\
\Pr(y_i=2)=&{\frac {e^{{\boldsymbol {\beta }}_{2}\cdot \mathbf x_i}}{1+\sum _{k=1}^{K-1}e^{{\boldsymbol {\beta }}_{k}\cdot \mathbf x_i}}}\\
&\cdots\\
\Pr(y_i=K-1)=&{\frac {e^{{\boldsymbol {\beta }}_{K-1}\cdot \mathbf x_i}}{1+\sum _{k=1}^{K-1}e^{{\boldsymbol {\beta }}_{k}\cdot \mathbf x_i}}}
\end{aligned}
\end{equation}
Next, for each observation, it uses the estimated logistic relations to predict the probability of cluster membership in all clusters, and the probabilities are then ordered in descending order; the vector of ordered probabilities can be indicated as $P_i = \left\{p_k^1,p_k^2,\ldots,p_k^K\right\}$. Consider a situation where $P_i = \left\{p_3^1=0.5,p_1^2=0.49,p_2^3=0.01\right\}$ (for $K=3$); while the highest probability is associated with $k=3$, it's not clear whether or not observation $i$ truly belongs in cluster $k=3$ or $k=1$. Since observation $i$ is clearly almost equidistant between both cluster means, it may not make sense to use it to recompute the mean of its assigned cluster $k=3$.

Our algorithm computes the ratio of the two largest of these probabilities:
\begin{equation}
R_i = \frac{p_k^1}{p_k^2}.
\end{equation}
The $i$th observation is only used to recompute the mean of cluster $y_i$ if $R_i > 1.5$. We can annotate this condition as $I\left(R_i\right)$, which returns $1$ if the inequality is met, and $0$ otherwise. We use $1.5$ because it allows the algorithm to consider any observations with greater than a 60:40 split between the two most likely clusters as being firmly placed in its cluster. Any belonging probability spread out among the remaining clusters (for $K>2$) only makes the algorithm more lenient. After augmentation in this manner, the The full Augmented k-means algorithm is:
\begin{enumerate}
\item For each observation $\mathbf{x_i}$ and cluster $k$, compute the squared Euclidean distance to the mean $\mathbf{C_k}$:
\begin{equation}
d_{ik} = \parallel\mathbf{x_i}-\mathbf{C_k}\parallel^2_2.
\end{equation}

\item Assign each observation to its closest cluster:
\begin{equation}
y_i = \underset{k=1,\ldots,K}{\argmin}d_{ik}.
\end{equation}

\item Perform logistic regression, with $\mathbf{x_i}$ as the independent data, and $y_i$ as the known class labels, then compute the cluster belonging probabilities, order them in descending order, and compute the ratio of the two largest probabilities $R_i=p_k^1/p_k^2$.

\item Recompute the cluster means:
\begin{equation}
\mathbf{C_k} = \frac{1}{n_k}\sum_{i=1}^n I\left(i,k\right) I\left(R_i\right)\mathbf{x_i},\ 
n_k = \sum_{i=1}^n I\left(i,k\right) I\left(R_i\right)
\end{equation}

\item Compute the total sum of squared distances:
\begin{equation}
S_t = \sum_{k=1}^K\sum_{i=1}^n I\left(i,k\right)d_{ik}.
\end{equation}

\item Measure the change in the total sum of squared distances from the previous iteration, and compare against $\epsilon$. If $\left\vert S_{t-1} - S_{t} \right\vert<\epsilon$, exit. Otherwise, return to step (i).
\end{enumerate}
Note that there's no $I\left(R_i\right)$ in the step (v); the sequence of total sum of squared distances does not take into account the additional information generated by the logistic regression. This computation is left unmolested so as to retain the algorithms property of monotonic convergence.

When Augmented k-means converges to a solution, every observation is classed into a cluster. However, the knowledge that certain observations had $I\left(R_i\right)=0$ is retained. Hence, in answer to the concerns addressed in \cite{MaitraRamler2009,TsengWong2005,Wong1982}, we could instead mark these observations as scatter.

\begin{figure}[H]
\centering
\subfloat[k-means: Class. Rate: 68.7\%, Iterations: 18 ]{\label{figdemo_K}\includegraphics[width=0.45\textwidth]{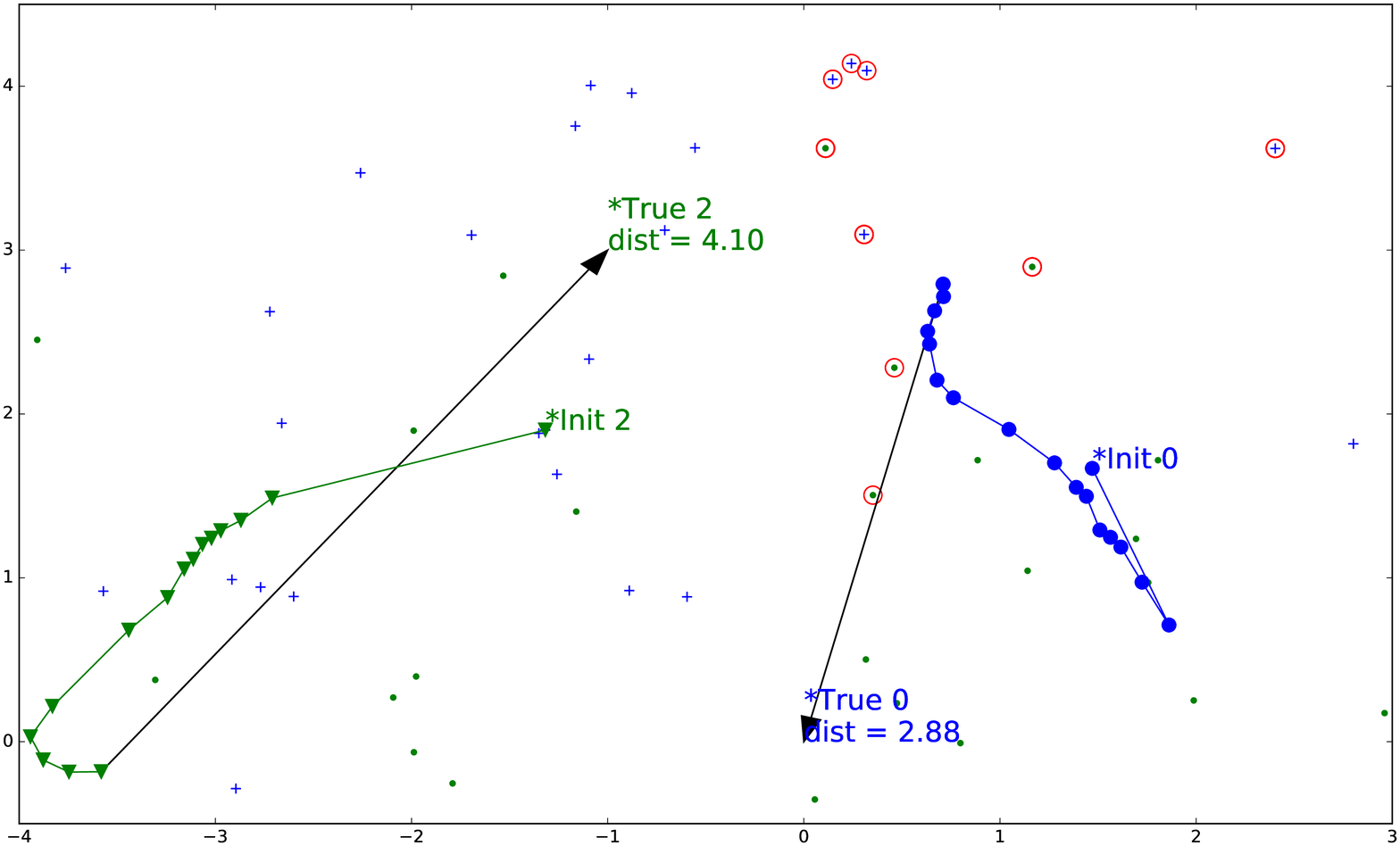}}
\hspace{0.1in}
\subfloat[Augmented k-means: Class. Rate: 74.7\%, Iterations: 7 ]{\label{figdemo_A}\includegraphics[width=0.45\textwidth]{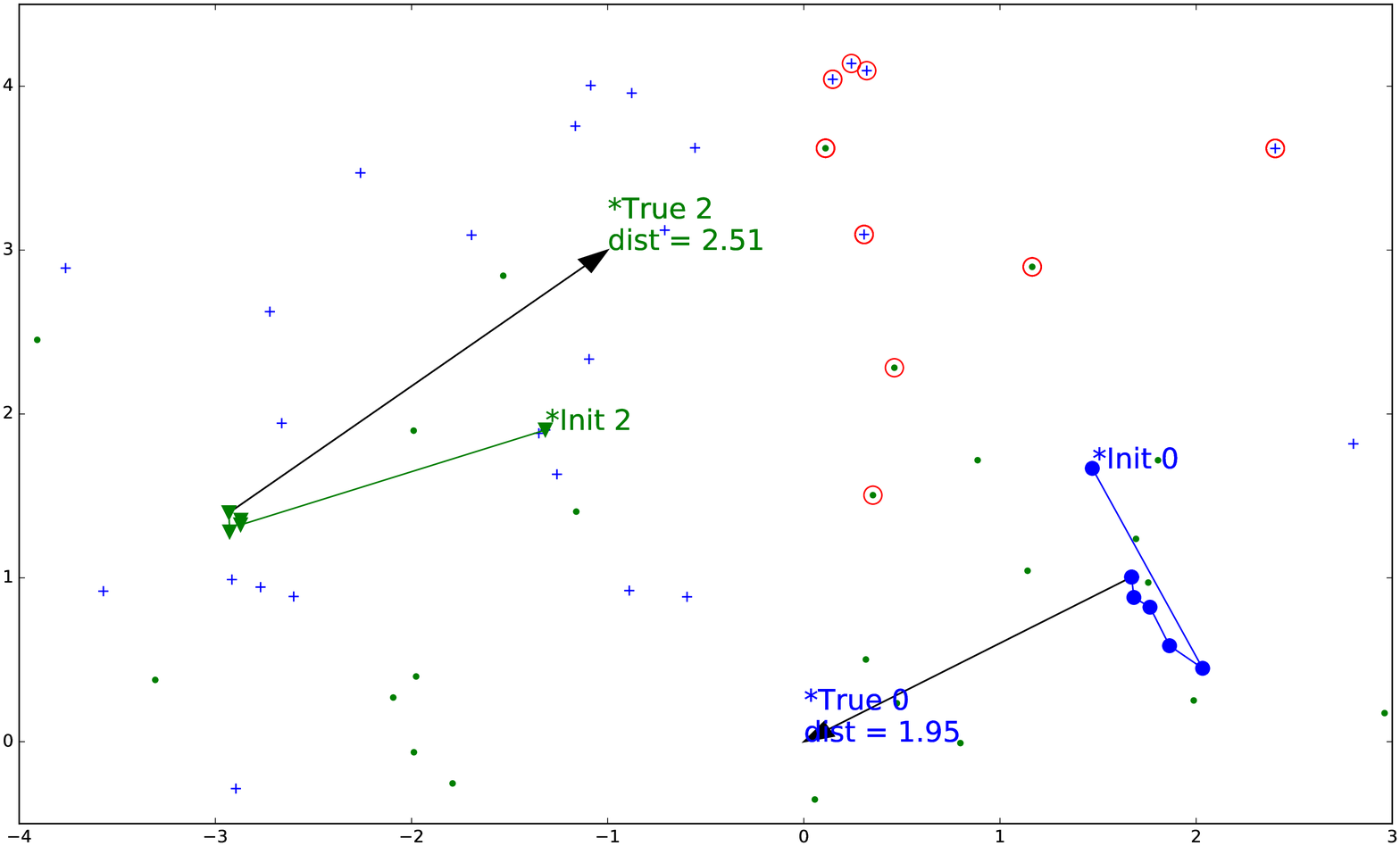}}
\caption{Evolution of Two Cluster Means, Contrasting k-means with Augmented k-means.}\label{figdemo}
\end{figure}

The two panes of Figure~\ref{figdemo} - generated from simulated data with overlapping clusters - help explain why Augmenting k-means is an improvement.
The solid blue circles in each pane shows how the mean of cluster 0 changed as the algorithm iterated, and the solid green triangles shows the same for cluster 2. The initial estimated cluster means from k-means++ are indicated with the "Init 0" / "Init 2" text, and the true cluster means are annotated with "True 0" / "True 2". In both panes, the datapoints are the small points, and those circled in red are the observations excluded by the augmentation from updating the cluster means.

In Figure~\ref{figdemo_K}, we see that the cluster means kept moving further away from the true centers. In Figure~\ref{figdemo_A}, however, we see that they stopped moving away after only a handful of iterations. The predominance of excluded observations in the upper right corner of the plots shows the reason. With k-means, these observations pulled the mean for cluster 0 in that direction. While not shown to make the plot more legible, there is an observation in the lower left corner which acted to pull the mean for cluster 2 in that direction. In each pane of Figure~\ref{figdemo}, the distance between the final and true cluster means are written near the true means. For both clusters, the final means computed by Augmented k-means are closer. It should be clear that the benefit obtained by the augmentation is very dependent on the homogeneity and scatter in the data, as well as the number of clusters and their spacing.

\section{\label{sec_results}Numerical Results}
Here we show comparative results for several datasets with known clustering structures. In both real data examples, we executed 1,000 replications of the algorithms, using k-means++ initialization. In each replication, both k-means and Augmented k-means began with the same initial state.

\subsection{\label{sec_simres}Simulated Data}
\begin{figure}[H]
\centering
\makebox{
\includegraphics[width=0.6\textwidth]{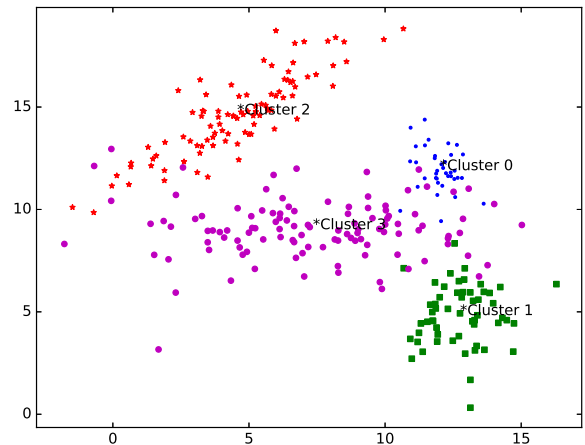}}
\caption{\label{figsim}Simulated Bivariate Data with Four Clusters}
\end{figure}
We begin with demonstrating the performance on a simulated bivariate dataset with $n=300$ observations from $K=4$ overlapping clusters. As can be seen in Figure~\ref{figsim}, there are a lot of observations that we can expect to not be firmly placed in a specific cluster. Instead of 1,000 replications, we ran 5,000, since the data was simulated. Each time, the same data with the same initial means was used for both algorithms. As shown in Table~\ref{tab_simd}, Augmented K-means correctly classified more observations than k-means $82\%$ of the time, and it only under-performed in $13\%$ of the replications. When it did outperform k-means, the classification gain was slightly more than $5\%$. Augmented k-means converged in fewer iterations in approximately half the replications; when it converged faster, it required $4.86$ fewer iterations on average. Averaged over all 5,000 replications, Augmented k-means only needed to run $0.55$s longer.
\begin{table}[H]
\caption{\label{tab_simd}Simulated Data: Frequency with which Augmented k-means Performance, Relative to k-means is:}
\centering
\fbox{
\begin{tabular}{rcc}
& Correct Class. Rate & Number Iterations\\\hline
Better & $81.8\%$ & $50.7\%$\\
Better or Equal & $86.5\%$ & $60.5\%$\\\hline
\multicolumn{3}{c}{Average Improvement When Augmented k-means is Better}\\
&$5.1\%$ & 4.86
\end{tabular}}
\end{table}

\subsection{\label{sec_irisres}Iris Data}
We continue with Fisher's iris data. This dataset consists of $p=4$ flower characteristics: \textit{petal length}, \textit{petal width}, \textit{sepal length}, and \textit{sepal width}. There are $K=3$ groups: $50$ observations each from the varieties \textit{Iris Setosa}, \textit{Iris Versicolor}, and \textit{Iris Virginica}. The comparative results for the Iris data are shown in Table~\ref{tab_iris}. In $95\%$ of the simulations, Augmented k-means correctly classified more observations, with an average improvement of $3.2\%$. For this dataset, the augmentation algorithm tended to require more iterations; in the $30\%$ of simulations in which it converged faster, Augmented k-means required on average 4.6 fewer iterations. For the rare simulations in which the classification performance was worse, the average shortfall relative to k-means was only $0.7\%$ - a single observation.
\begin{table}[H]
\caption{\label{tab_iris}Iris Data: Frequency with which Augmented k-means Performance, Relative to k-means is:}
\centering
\fbox{
\begin{tabular}{rcc}
& Correct Class. Rate & Number Iterations\\\hline
Better & $95.3\%$ & $31.3\%$\\
Better or Equal & $99.9\%$ & $35.1\%$\\\hline
\multicolumn{3}{c}{Average Improvement When Augmented k-means is Better}\\
&$3.2\%$ & 4.59
\end{tabular}}
\end{table}

\subsection{\label{sec_wineres}Wine Composition Data}
Our final example is the wine recognition dataset of M. Fiorina, \textit{et al.}, used in \cite{AeberhardCoomansEtal1992}. These data are the results of a chemical analysis of $n=178$ wines grown in the same region in Italy but derived from $K=3$ different cultivars ($n_1=59$, $n_2=71$, $n_3=48$). The analysis determined the values of $p=13$ characteristics of each wine. The variables are shown in Table~\ref{tab_winedata}.
\begin{table}[H]
\caption{\label{tab_winedata}Wine Data Variables.}
\centering
\fbox{%
\begin{tabular}{rl|rl}
\multicolumn{2}{c|}{Variable} & \multicolumn{2}{|c}{Variable}\\\hline
$x_1$ & Alcohol & $x_8$ & Non-flavonoid Phenols\\
$x_2$ & Malic Acid & $x_9$ & Proanthocyanins\\
$x_3$ & Ash & $x_{10}$ & Color Intensity\\
$x_4$ & Alcalinity of Ash & $x_{11}$ & Hue\\
$x_5$ & Magnesium & $x_{12}$ & OD280/OD315 of Diluted Wines\\
$x_6$ & Total Phenols & $x_{13}$ & Proline\\
$x_7$ & Total Flavonoids & &
\end{tabular}}
\end{table}
For the wine data, Augmented k-means outperformed k-means in $78\%$ of the simulations, regarding classification, and $59\%$ regarding iteration count, as can be seen in Table~\ref{tab_wine}. While the improvement in classification performance was slight, the average additional computation time required by Augmented k-means was only $0.11$s. When it needed more iterations than k-means, the excess was less than 2 iterations on average.
\begin{table}[H]
\caption{\label{tab_wine}Wine Data: Frequency with which Augmented k-means Performance, Relative to k-means is:}
\centering
\fbox{
\begin{tabular}{rcc}
& Correct Class. Rate & Number Iterations\\\hline
Better & $78.2\%$ & $59.2\%$\\
Better or Equal & $83.0\%$ & $84.0\%$\\\hline
\multicolumn{3}{c}{Average Improvement When Augmented k-means is Better}\\
&$0.7\%$ & 4.59
\end{tabular}}
\end{table}

\section{\label{sec_conclude}Concluding Remarks}
In this technical report, we've developed a new clustering algorithm, called Augmented k-means, that combines unsupervised clustering with k-means and supervised clustering with logistic regression. In each iteration, we use the group membership probabilities from logistic regression to exclude observations used to recompute the cluster means in k-means. This allows each cluster to form without being influenced by observations which don't firmly belong. We have demonstrated the advantages of Augmented k-means on both simulated and real datasets. The augmentation frequently leads to better classification performance and / or faster convergence.

It is true that our results demonstrate minimal incremental performance improvement over k-means++, which could be seen as a reason to forego publication. However, we feel that our hybrid unsupervised + supervised clustering approach is sufficiently innovative to justify publication.  Additional work around this innovation will only be accelerated by sharing the idea openly in the research community.

Further research with Augmented k-means could go in a few directions. The most obvious would be to attempt to augment k-means with a different supervised learning procedure, such as discriminant analysis (linear, quadratic, or kernel) or artificial neural networks. Of course, with both these procedures, the researcher has several subjective parameterization decisions to make. Also, neither produces the cluster belonging probabilities, so some other model output would need to be used in their place. It could be worthwhile to include a feature selection procedure in the logistic regression step in our algorithm. While this would require more CPU time, modeling the predictive power in an optimal subset should cause more greedy exclusion of observations, which may further improve performance.

\bibliography{AugClust}
\bibliographystyle{elsart-harv}
\end{document}